\documentclass{article}

\PassOptionsToPackage{numbers, compress}{natbib}

\usepackage[final]{neurips_2024}

\usepackage[utf8]{inputenc} 
\usepackage[T1]{fontenc}    
\usepackage{hyperref}
\usepackage{url}            
\usepackage{booktabs}       
\usepackage{amsfonts}       
\usepackage{nicefrac}       
\usepackage{microtype}      
\usepackage{xcolor}         
\newcommand{\method}[1]{{ImOV3D}}
\usepackage{subfiles}
\usepackage{caption}
\usepackage{graphicx}
\usepackage{wrapfig}
\usepackage{floatflt}
\usepackage{amsmath}
\usepackage{amssymb}
\usepackage{booktabs}
\usepackage{float}
\usepackage{lipsum}
\usepackage{afterpage}
\usepackage{multirow}
\usepackage{colortbl}
\usepackage{hhline}
\usepackage{adjustbox}
\usepackage{tabularx}
\usepackage{siunitx}
\usepackage{pifont}
\usepackage{tikz}
\usepackage{bbding}
\usepackage{pifont}
\usepackage{makecell}
\usepackage[accsupp]{axessibility}  
\usepackage{orcidlink}
\usepackage{amsmath}
\usepackage{adjustbox}
\usepackage{cleveref} 

\crefname{section}{Sec.}{Secs.}
\Crefname{section}{Section}{Sections}
\Crefname{table}{Table}{Tables}
\crefname{table}{Tab.}{Tabs.}
\newsavebox{\mycustomt}
\newsavebox{\mycustomb}
\newsavebox{\mycustomr}
\definecolor{babyblue}{rgb}{0.92,0.96,1.0}
\definecolor{red}{rgb}{1.0,0.65,0.0}
\definecolor{rred}{rgb}{1.0,0.0,0.0}
\definecolor{ggreen}{rgb}{0.0,0.8,0.0}
\definecolor{brightred}{rgb}{1.0,0.0,0.0}
\definecolor{brightpurple}{rgb}{0.75, 0.0, 1.0}
\definecolor{brightgreen}{rgb}{0.0, 0.6, 0.0}
\definecolor{brightblue}{rgb}{0.0, 0.6, 1.0}
\definecolor{green}{rgb}{0.80, 0.98, 0.80}
\definecolor{orange}{rgb}{1.0, 0.93, 0.83}
\definecolor{yellow}{rgb}{0.97,0.98,0.86}
\definecolor{gray}{rgb}{0.5,0.5,0.5}
\definecolor{pink}{rgb}{1, 0.94, 0.96}
\definecolor{purple}{rgb}{0.97, 0.93, 1.0}
\definecolor{myblue}{rgb}{0.8, 0.8, 1}

\newcommand{\pink}{\cellcolor{pink}}
\newcommand{\purple}{\cellcolor{purple}}
\newcommand{\green}{\cellcolor{green}}
\definecolor{deepblue}{rgb}{0.0, 0.0, 0.9}

\author{
  Timing Yang$^{1,2}$\thanks{Equal contribution.} \quad
  Yuanliang Ju$^{1,2}$\footnotemark[1] \quad
  Li Yi$^{2,3,1}$\thanks{Corresponding author.} \quad \\
  $^1$ Shanghai Qi Zhi Institute,  
  $^2$ IIIS, Tsinghua University,
  $^3$ Shanghai AI Lab    \\
}

\title{ImOV3D: Learning \underline{O}pen-\underline{V}ocabulary Point Clouds \underline{3D} Object Detection from Only 2D \underline{Im}ages}

\begin{document}

\maketitle
\begin{abstract}
    Open-vocabulary 3D object detection (OV-3Det) aims to generalize beyond the limited number of base categories labeled during the training phase. The biggest bottleneck is the scarcity of annotated 3D data, whereas 2D image datasets are abundant and richly annotated. Consequently, it is intuitive to leverage the wealth of annotations in 2D images to alleviate the inherent data scarcity in OV-3Det. In this paper, we push the task setup to its limits by exploring the potential of using solely 2D images to learn OV-3Det. The major challenges for this setup is the modality gap between training images and testing point clouds, which prevents effective integration of 2D knowledge into OV-3Det. To address this challenge, we propose a novel framework \textbf{ImOV3D} to leverage pseudo multimodal representation containing both images and point clouds (PC) to close the modality gap. The key of ImOV3D lies in flexible modality conversion where 2D images can be lifted into 3D using monocular depth estimation and can also be derived from 3D scenes through rendering. This allows unifying both training images and testing point clouds into a common image-PC representation, encompassing a wealth of 2D semantic information and also incorporating the depth and structural characteristics of 3D spatial data. We carefully conduct such conversion to minimize the domain gap between training and test cases. Extensive experiments on two benchmark datasets, SUNRGBD and ScanNet, show that ImOV3D significantly outperforms existing methods, even in the absence of ground truth 3D training data. With the inclusion of a minimal amount of real 3D data for fine-tuning, the performance also significantly surpasses previous state-of-the-art. Codes and pre-trained models are released on the  \url{https://github.com/yangtiming/ImOV3D}.
    
\end{abstract}

\section{Introduction}
\label{sec:intro}
In the 3D vision community, there is a notable surge in interest surrounding open-vocabulary 3D object detection (OV-3Det). This task focuses on the detection of objects from unbounded categories that were not present during the training phase, using 3D point clouds as input. Such capability holds immense significance in dynamic 3D environments where a wide range of object categories constantly emerge and evolve, which is critical in downstream applications including robotics\cite{chen2023open, conceptgraphs, lu2023ovir, ju2024robo}, autonomous driving \cite{ma2022rethinking,wang2024omnidrive}, and augmented reality \cite{nuernberger2016snaptoreality,wagner2009real}. 

\begin{figure}[t]
    \centering
    \includegraphics[width=1\linewidth]{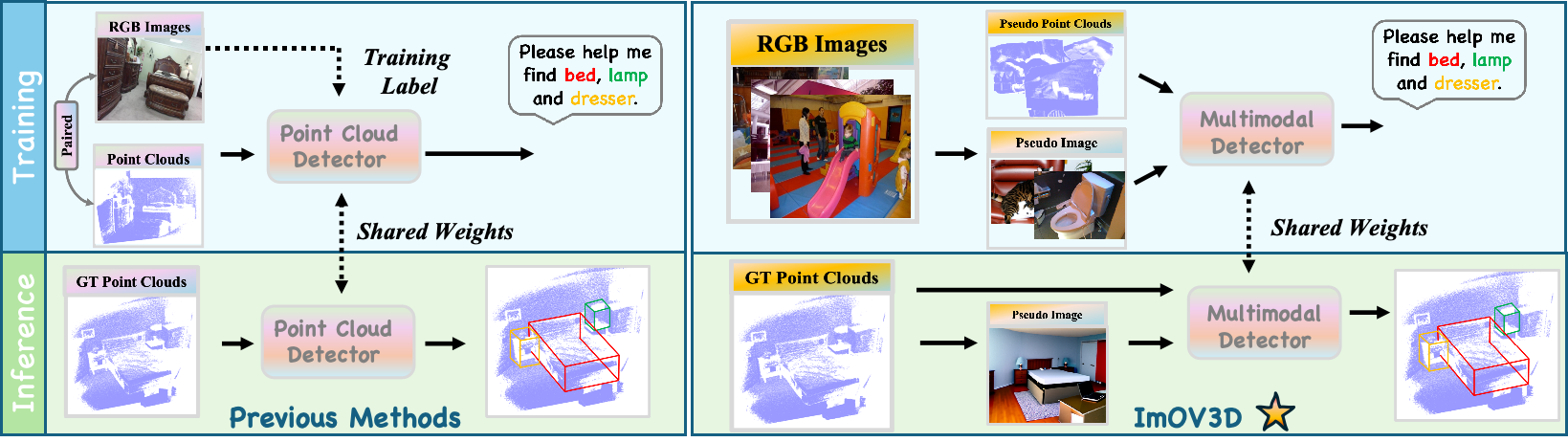}
    \caption{\textbf{Left:} Traditional methods require paired RGB-D data for training and use single-modality point clouds as input during inference. \textbf{Right:} \method{} involves using a vast amount of 2D images to generate pseudo point clouds during the training phase, which are then rendered back into images. In the inference phase, with only point clouds as input, we still construct a pseudo-multimodal representation to enhance detection performance.}
    \label{fig:test1}
    \vspace{-0.5cm}
\end{figure}

With the advancements in OV-3Det, which is not only scarce in terms of labels but also in the data itself. However, the collection and annotation of 3D point clouds scenes pose significant challenges. The availability of accessible and scannable scenes \textit{(e.g. indoor scenes)} may be limited. Additionally, obtaining 3D annotations often requires substantial human effort and time-consuming. These limitations impact the model's performance in handling novel objects. Existing methods \cite{ov3det,OV-3DETIC,lu2023ovir,chen2024towards,yu2022data} seek help from powerful open-vocabulary 2D detectors. A common method leverages paired RGB-D data together with 2D detectors to generate 3D pseudo labels to address the label scarcity issue, as shown in Figure \ref{fig:test1} left. But they are still restricted by the small scale of existing paired RGB-D data. Moreover, the from scratch trained 3D detector can hardly inherit from powerful open-vocabulary 2D detector models directly due to the modality difference. We then ask the question, what is the best way to transfer 2D knowledge to 3D for OV-3Det?

Observing that the modality gap prevents a direct knowledge transfer, we propose to leverage a pseudo multi-modal representation to close the gap. On one hand, we can lift a 2D image into a pseudo-3D representation through estimating the depth and camera matrix. On the other hand, we can convert a 3D point cloud into a pseudo-2D representation through rendering. The pseudo RGB image-PC multimodal representation could serve as a common ground for better transferring knowledge from 2D to 3D.

In this paper, we present \method{}, which addresses these challenges by employing pseudo-multimodal representation as a unified framework. As shown in Figure \ref{fig:test1} right side, In both the training and the inference phase, we construct pseudo-multimodal representation to achieve our goal of training solely with 2D images and better integrating multimodal features to enhance the performance of OV-3Det. Our key idea lies in proper modality conversion. Specifically, the entire pipeline consists of two flows: (1) \textbf{Image $\rightarrow$ Pseudo PC}, by leveraging a large-scale 2D images training set, our method begins by converting images to pseudo point clouds through monocular depth estimation and approximate camera parameter. We automatically generate pseudo 3D labels based on 2D annotations, providing the necessary training data. We also designed a set of revision modules, which significantly improve the quality of the pseudo 3D data through the use of GPT-4 \cite{achiam2023gpt}'s size prior and the orientation of the estimated normal map. (2) \textbf{Pseudo PC $\rightarrow$ Pseudo Image}, we learn a point clouds renderer capable of producing natural-looking textured 2D images from pseudo 3D point clouds. This enables \method{} to leverage pseudo-multimodal 3D detection even for point cloud-only inputs during inference, transferring 2D rich semantic information and proposals into the 3D space, further enhancing the detector's performance.

Despite being trained solely with the 2D image set, \method{} exhibits impressive detection results when directly processing real 3D scans. This is attributed to the high fidelity of the lifted point clouds and the point clouds rendering. Additionally, when a small amount of real 3D data becomes available, even without any 3D annotations, \method{} can further narrow the gap between pseudo and real data by fine-tuning on such 3D data, leading to improved detection performance. To validate the effectiveness of \method{}, we perform extensive experiments on two benchmark datasets: SUNRGBD \cite{sunrgbd} and ScanNet \cite{scannet}. Notably, in scenarios where real 3D training data is unavailable, \method{} surpasses previous state-of-the-art open-vocabulary 3D detectors by an mAP@0.25 improvement of at least 7.14\% on SUNRGBD and 6.78\% on ScanNet. Furthermore, when real 3D training data is accessible, \method{} continues to outperform various challenging baselines by a large margin. Thorough ablations are also conducted to validate the efficacy of our designs. In summary, our contributions are three-fold:
\begin{itemize}
    \item We propose \method{}, the first OV-3Det method that can be trained solely with 2D images \textbf{without requiring any 3D point clouds or 3D annotations}.
    \item We introduce a novel \textbf{pseudo-multimodal representation} pipeline which converts 2D internet images and corresponding detections into pseudo point clouds, pseudo 3D annotations, and point clouds renderings to support point clouds-based multimodal OV-3Det.
    \item \method{} achieves state-of-the-art performance on two general OV-3Det benchmark datasets across various settings, showcasing its ability to enhance open world 3D understanding despite the lack of 3D data and annotations.
\end{itemize}

\section{Related Work}
\par\noindent\textbf{Open-Vocabulary 2D Object Detection} encompasses two primary series of works: The first \cite{OVcondition, yin, zsd-yolo,  edadet, bridging,vild,pham2024lp,wang2023object,cho2023open,chen2023ovarnet,OVcondition},  which draws upon knowledge from pre-trained Vision-Language models (e.g., CLIP \cite{CLIP}), comprehends the relationships between images and their corresponding textual descriptions, thereby enhancing object recognition and classification. The second series \cite{glip, detclip, detclipv2,caption,pseudo,regionclip,minderer2205simple,kim2023detection,detic} involves the use of extensive training data, specifically text/image pairs, enabling the model to learn a more diverse set of object representations. Detic \cite{detic} leverages vocabularies from image classification datasets to train the classification head of object detectors, addressing the issue of insufficient training data and enabling inference on a larger vocabulary set. In the 2D component of our pseudo- multimodal detector, we utilize Detic \cite{detic} to predict 2D labels and bounding boxes. These 2D visual information features are then converted and augmented for the 3D point cloud detector, significantly enhancing our model's ability to recognize a broader range of objects.

\par\noindent\textbf{Open-Vocabulary Scene Understanding} has recently gained increased attention \cite{openscene,clip-field,clipfo3d,clipgo3d,conceptfusion,lerf,pla,chen2024towards} and plays a critical role in robotics, autonomous driving, \textit{etc.} OpenScene \cite{openscene} achieves open-world scene understanding without the need for labeled data by densely embedding 3D scene points together with text and image pixels into the CLIP \cite{CLIP} feature space. PLA \cite{pla} develops a hierarchical approach to pairing 3D data with text for open-world 3D learning. We focus on OV-3Det, where merely extracting CLIP \cite{CLIP} features is insufficient. We also require the intricate spatial structure of point clouds to enhance detection accuracy and robustness. By integrating both CLIP \cite{CLIP}'s visual knowledge and the detailed geometric information from point clouds, our approach aims to enable the recognition of a broader range of objects beyond the predefined categories.

\par\noindent\textbf{Open-Vocabulary 3D Object Detection} in 3D vision is still in its early stages, especially when compared to traditional 3D object detection \cite{VoteNet, imvotenet, 3DETR, GroupFree3D,cao20243dgs}. OV-3DETIC \cite{OV-3DETIC} leverages ImageNet1K \cite{imagenet} to expand the detector's vocabulary set and conducts contrastive learning between images and point clouds modalities for more effective knowledge transfer. OV-3DET \cite{ov3det} generates pseudo 3D annotations for localization using a pre-trained 2D open-vocabulary detector \cite{detic}. CoDA \cite{coda} leverages 2D and 3D prior information and a cross-modal alignment module to simultaneously learn the localization and classification capabilities. CoDAv2 \cite{cao2024collaborative} improves CoDA \cite{coda} further by proposing the novel object enrichment strategy and 2D box guidance. FM-OV3D \cite{zhang2023fm} combines multiple foundation models without the need for 3D annotations. However, they are still subject to the influence of the volume of 3D data and still require strict correspondence between RGB-D data. Our method can generate training data for OV-3Det task using only 2D images, without any 3D ground truth data. It can directly achieve state-of-the-art performance when tested on the evaluation set. The designed pseudo-multimodal representation pipeline provides a novel solution for the utilization of both 2D and 3D information.

\begin{figure}[t]
    \centering
    \includegraphics[width=1.0\linewidth]{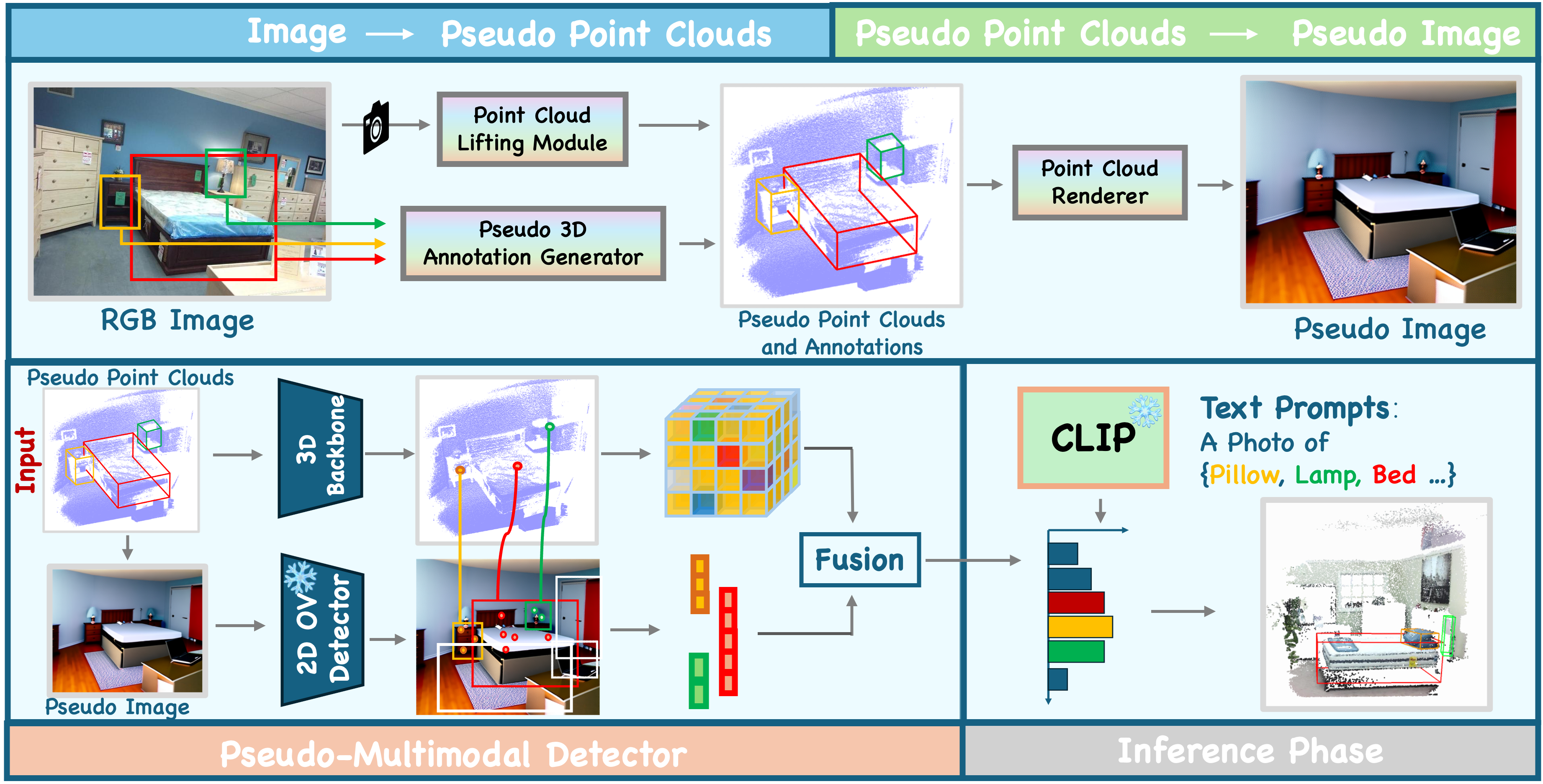}
    \caption{Overview of \method{}: Our model takes 2D images as input and puts them into the Pseudo 3D Annotation Generator to produce pseudo annotations. These 2D images are also fed into the Point Cloud Lifting Module to generate pseudo point clouds. Subsequently, using the Point Cloud Renderer, these pseudo point clouds are rendered into pseudo images, which then get processed by a 2D open vocabulary detector to detect 2D proposals and transfer the 2D semantic information to 3D space. Armed with pseudo point clouds, annotations, and pseudo images data, we proceed to train a multimodal 3D detector.}
    \label{fig:pipeline}
     \vspace{-0.5cm}
\end{figure}

\vspace{-0.4cm}
\section{Method}
\label{sec:method}

\vspace{-0.2cm}
\subsection{Overview}
An overview of the proposed open world 3D Object Detection model, \textbf{\method{}}, is shown in Figure \ref{fig:pipeline}. \method{} is a point cloud-only model that addresses the challenges of the scarcity of annotated 3D datasets in open-vocabulary 3D Object Detection. To overcome this challenge, \method{} uses large-scale 2D datasets to generate pseudo 3D point clouds and annotations. We use a monocular depth estimation model to create metric depth images, which are then converted into pseudo 3D point clouds for both indoor and outdoor scenes. To generate pseudo 3D annotations, we lift 2D bounding boxes into 3D space. To leverage multimodal data, we transform the point clouds into pseudo images using a point cloud renderer. Our training strategy involves a two-stage approach. Firstly, we conduct pre-training using pseudo 3D point clouds and corresponding annotations. Subsequently, we initiate an adaptation stage aimed at minimizing the domain discrepancy between 2D and 3D datasets.

\vspace{-0.2cm}
\subsection{Point Cloud Lifting Module}
\vspace{-0.2cm}

The success of open-vocabulary object detection relies heavily on the availability of large-scale labeled datasets. However, the scarcity of comparable 3D datasets poses a challenge for open world 3D Object Detection. To address this, we bridge 2D images $\mathcal{I}_{\text{2D}} \in \mathbb{R}^{M \times H \times W \times 3}$ (where $M$ is the number of images, and $H$ and $W$ are the height and width, respectively) to 3D detection by generating pseudo 3D point clouds $\mathcal{P}_{\text{pseudo}} \in \mathbb{R}^{M \times N \times 3}$ (where $N$ is the number of points, each with coordinates (x, y, z)).

Utilizing 2D datasets for 3D detection presents difficulties due to the absence of metric depth images and camera parameters. To overcome these obstacles, we use a metric depth estimation model to obtain single-view depth images $\mathcal{D}_{\text{metric}} \in \mathbb{R}^{M \times H \times W}$. Additionally, we employ fixed camera intrinsics $K \in \mathbb{R}^{3 \times 3}$, with the focal length \(f\) calculated based on a 55-degree field of view (FOV) and the image dimensions.

\begin{figure}[t]
    \centering
    \includegraphics[width=1\linewidth]{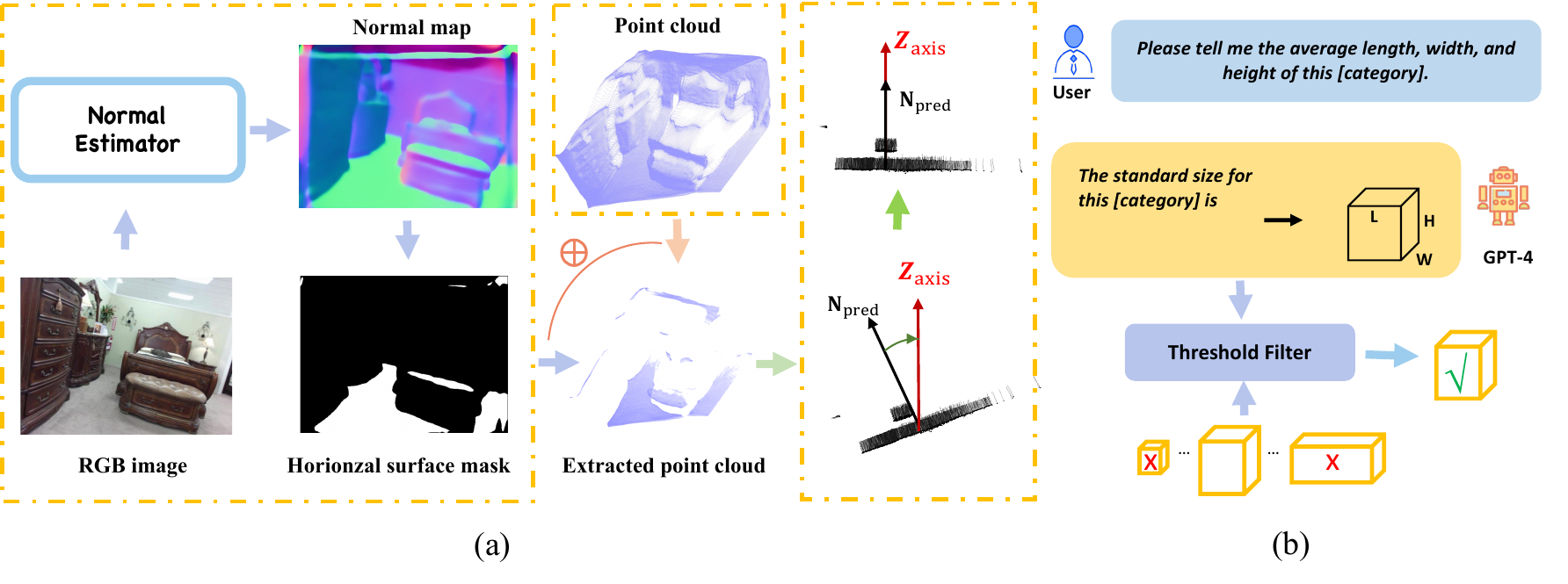}
    \caption{Illustration of 3D Data Revision Module: \textbf{(a)} The rotation correction module involves processing an RGB image through a Normal Estimator to generate a normal map. This map then helps extract a horizontal surface mask for identifying horizontal point clouds, from which normal vectors \(N_{pred}\) are obtained. These vectors are aligned with the Z-axis to compute the rotation matrix \(R\). \textbf{(b)} In the 3D box filtering module, prompts related to object dimensions are first provided to GPT-4 to determine the mean size for each category. This mean size is then used to filter out boxes that do not meet the threshold criteria.}
    \label{fig:HSANEandbox}
    \vspace{-0.5cm}
\end{figure}

However, the absence of camera extrinsics $\mathbf{E} = \{ R \mid t\}$ (where $R$ is the rotation matrix and $t$ is the translation vector set to $[0, 0, 0]^{\top}$) results in the arbitrary orientation of point clouds. To correct this, we use a rotation correction module to ensure the ground plane is horizontal, as shown in Figure \ref{fig:HSANEandbox} (a).  First, we estimate the surface normal vector at each pixel using a normal estimation model \cite{bae2021estimating}, creating a normal map. From this, we selectively extract the horizontal normal vector ${N}_i$ at each pixel, defined as $\left(N_x, N_y, N_z\right)$. We then compute the normal vector of the horizon surface as ${N}_{\text{pred}} = Cluster({N}_i)$. To align ${N}_{\text{pred}}$ with the $Z_{axis}$, we calculate the rotation matrix $R$ using the following equation:
\vspace{-0.2cm}
\begin{equation}\label{eq:HSANE1}
\small
R=I+K+K^2 \frac{1-N_{pred} \cdot Z_{axis}}{\|v\|^2}
\end{equation}

where $I$ is the identity matrix, $v$ is the cross product of $N_{pred}$ and $Z_{axis}$, expressed as $N_{pred} \times Z_{axis}$, $K$ is the skew symmetric matrix constructed from the vector $v$, represented as: 
\begin{equation}\label{eq:HSANE2}
K=\left[\begin{array}{ccc}
0 & -v_z & v_y \\
v_z & 0 & -v_x \\
-v_y & v_x & 0
\end{array}\right]
\end{equation}

After obtaining the camera intrinsics matrix \( K \) and the camera extrinsics matrix \( \mathbf{E} \) through the previous steps, depth images $\mathcal{D}_{\text{metric}}$ are converted into point clouds \(\mathcal{P}_{\text{pseudo}}\).


\vspace{-0.2cm}
\subsection{Pseudo 3D Annotation Generator}
\vspace{-0.2cm}

 Building upon the vast collection of pseudo 3D point clouds $\mathcal{P}_{\text{pseudo}}$ acquired from 2D datasets, our next step is to generate pseudo 3D bounding boxes $\mathcal{B}_{\text{3Dpseudo}} \in \mathbb{R}^{M \times K \times 7}$(where \(K\) is the number of bounding boxes and each box has 7 parameters: center coordinates, dimensions, and orientation).

2D datasets contain rich segmentation information that can be used to generate 3D boxes by lifting. Using the camera intrinsics matrix \( K \) and camera extrinsics matrix \( \mathbf{E} \) obtained through the Point Cloud Lifting Module, we lift the 2D bounding boxes $\mathcal{B}_{2Dgt} \in \mathbb{R}^{M \times K \times 4}$ from the 2D datasets into 3D space by extracting 3D points that fall within the predicted 2D boxes, generating frustum 3D boxes $\mathcal{B}_{\text{3Dpseudo}}$. The extracted point clouds may contain background points and outliers. To remove these, we employ a clustering \cite{DBscan} algorithm to analyze point clouds. Through the clustering results, we can identify and remove background points and outliers that do not belong to the target objects.

However, these lifted 3D boxes may still contain noise from the depth images $\mathcal{D}_{\text{metric}}$ obtained by monocular depth estimation. To address this issue, we use a 3D box filtering module to filter out inaccurate 3D boxes, as shown in Figure \ref{fig:HSANEandbox} (b). First, we construct a database of median object sizes using GPT-4 \cite{achiam2023gpt}. By prompting GPT-4 with "Please tell me the average length, width, and height of this [category], using meters as the unit", we obtain the median dimensions $L_{GPT}, W_{GPT}, H_{GPT}$. Each object in a scene, defined by dimensions $L, W, H$, is compared to these median dimensions using a threshold $T$. An object is preserved if each element of:

\vspace{-0.5cm}
\begin{equation}
\small
\label{eq:3DBoxSanitizerCriteria}
T < \frac{R}{R_{\text{GPT}}} < \frac{1}{T}, \quad \forall R \in \{L, W, H\}
\end{equation}



The 3D box filtering module consists of two components: Train Phase Prior Size Filtering and Inference Phase Semantic Size Filtering. The first component filters out boxes that do not match the size criteria before training. The second component removes semantically similar but size-different categories during inference, preventing errors such as misidentifying a book as a bookcase.

\vspace{-0.2cm}
\subsection{Point Cloud Renderer}
\vspace{-0.2cm}

Point cloud data has inherent limitations, such as the inability of sparse point clouds to capture detailed textures. 2D images can enrich 3D data by providing additional texture information that point clouds lack. To utilize 2D images, we transform point clouds $\mathcal{P}$ into rendered images $\mathcal{I}_{\text{rendered}} \in \mathbb{R}^{M \times H \times W}$.

Integrating rendered images into a 3D detection pipeline is challenging. A naive approach, as mentioned in PointClip \cite{pointclipv2}, is to append raw depth values across the RGB channels, but this fails to apply a mature open-world 2D detector effectively. To leverage multimodal information without additional inputs beyond 3D point clouds, we develop a point cloud renderer to convert point clouds into detailed pseudo images. This process can also be learned solely from 2D image datasets.

The point cloud renderer has two key modules: The point cloud rendering module converts point clouds $\mathcal{P}$ into rendered images $\mathcal{I}_{\text{rendered}}$, and the color rendering module then processes these images to produce colorized outputs using ControlNet \cite{zhang2023adding}. ControlNet \cite{zhang2023adding} is a method designed to control diffusion models, transforming rendered images $\mathcal{I}_{\text{rendered}}$ into pseudo images $\mathcal{I}_{\text{pseudo}} \in \mathbb{R}^{M \times H \times W \times 3}$.

In the pretraining stage, we use the camera intrinsics \( K \) and extrinsics \( \mathbf{E} \) from the Point Cloud Lifting Module to render $\mathcal{P}_{\text{pseudo}}$ into rendered images $\mathcal{I}_{\text{rendered}}$. During adaptation and inference, we render ground truth point clouds $\mathcal{P}_{\text{gt}}$ into images using the intrinsics \( K \) obtained in the same way. Due to the lack of extrinsics \( \mathbf{E} \), the final rendered images $\mathcal{I}_{\text{rendered}}$ are obtained by finding the optimal angle from different horizontal and vertical perspectives to make the images most compact.

In reality, we cannot project a point cloud while guaranteeing that every pixel corresponds to some points. There will be holes and missing areas due to point cloud imperfections or incompatible viewpoint selection. We adjust the camera's position horizontally and vertically to observe point clouds from various angles, removing obscured portions. Finally, we render the point clouds back into images from their original perspective, resulting in partial view rendered images $\mathcal{I}_{\text{partial}} \in \mathbb{R}^{M \times N \times 3}$. The angle range for adjustments is set from -75 to 75 degrees, with a 15-degree interval:
\vspace{-0.1cm}
\begin{equation}\label{eq:Render}
\theta_h, \theta_v \in \{-75 + 15k | k = 0, 1, 2, \ldots, 10\}^\circ
\end{equation}
\vspace{-0.1cm}
where \(k\) is an integer indicating the stepwise adjustment of the camera's angle.

After generating partial view rendered images $\mathcal{I}_{\text{partial}}$, the next step is to fine-tune ControlNet \cite{zhang2023adding} using these images to obtain pseudo images $\mathcal{I}_{\text{pseudo}}$. Three types of data are prepared for fine-tuning: prompts, targets, and sources. RGB images $\mathcal{I}_{\text{2D}}$ from a 2D dataset serve as the targets, while the partial view rendered images $\mathcal{I}_{\text{partial}}$ are the training sources. Prompts are not used during training.

Finally, we use the pseudo images $\mathcal{I}_{\text{pseudo}}$ and annotations $\mathcal{B}_{\text{2Dgt}}$ from 2D datasets to fine-tune an open-vocabulary 2D detector. Thus, we can use $\mathcal{I}_{\text{pseudo}}$ to obtain corresponding pseudo 2D bounding boxes $\mathcal{B}_{\text{2DTpseudo}} \in \mathbb{R}^{M \times K \times 4}$.

\vspace{-0.2cm}
\subsection{Pseudo Multimodal 3D Object Detector}
\vspace{-0.2cm}

With an extensive dataset comprising abundant 3D data ($\mathcal{P}$ + $\mathcal{B}_{\text{3Dpseudo}}$) and pseudo images data $\mathcal{I}_{\text{pseudo}}$, our next step is to train a pseudo multimodal 3D detector using a two-stage approach.

\textbf{Training Strategy}
Our training process includes pretraining and adaptation stages. In the pretraining stage, we train on pseudo 3D point clouds $\mathcal{P}_{\text{pseudo}}$ and annotations $\mathcal{B}_{\text{3Dpseudo}}$, combined with pseudo images $\mathcal{I}_{\text{pseudo}}$. While the pre-trained model performs well for zero-shot detection, a significant domain gap exists between 2D and 3D datasets. 

In the adaptation stage, to minimize the domain gap, we follow the same approach as OV-3DET. First, a pre-trained open-vocabulary 2D detector is used to detect objects in the image. Then, these 2D boxes $\mathcal{B}_{\text{2Dpseudo}} \in \mathbb{R}^{M \times K \times 4}$, along with RGBD data, are lifted into 3D space. Through clustering to remove background and outlier points, we obtain precise and compact 3D boxes $\mathcal{B}_{\text{3Dpseudo}}$. Finally, this processed data is used for adaptation. To further explore the benefits of pretrain, we use 3D datasets of varying sizes to test the model's performance under different data availability conditions.

\textbf{Loss Function}
In this section, we describe loss function used in the pretrain stage. By leveraging $\mathcal{P}_{\text{pseudo}}$ and \(\mathcal{B}_{3D\text{pseudo}}\), a 3D backbone is trained to obtain seed points \(\mathcal{K} \in \mathbb{R}^{K \times 3}\), where \(K\) represents the number of seeds, along with 3D feature representations \(F_{pc} \in \mathbb{R}^{K \times (3 + F)}\), with \(F\) denoting the feature dimension. Then, seed points are projected back into 2D space via the camera matrix. These seeds that fall within the 2D bounding boxes $\mathcal{B}_{2D\text{Tpseudo}}$ retrieve the corresponding 2D cues associated with these boxes and bring them back into 3D space. These lifted 2D cues features are represented as \(F_{img} \in \mathbb{R}^{K \times (3 + F')}\), where \(F'\) represents the feature dimension. Finally, the point cloud features \(F_{pc}\) and image features \(F_{img}\) are concatenated, forming the joint representation \(F_{joint} \in \mathbb{R}^{K \times (3+F+F')}\). In the adaptation stage, \(\mathcal{P}_{\text{pseudo}}\) is replaced with \(\mathcal{P}_{\text{gt}}\), keeping the workflow consistent with the pretrain stage.

\vspace{-0.5cm}

\begin{equation}
\small
\label{eq:loss}
\mathcal{L}_{\text{total}} = \mathcal{L}_{\text{loc}} +  \sum_{i} W_i \times \text{CrossEntropy}(\text{Cls-header}(\mathcal{F}_{i}) \cdot \mathcal{F}_{\text{text}}) 
\end{equation}
\vspace{-0.5cm}

where \( i \) represents different features, such as \( \text{pc} \), \( \text{img} \), \( \text{joint} \). \( W_i \) is the weight corresponding to feature \( i \). \( \mathcal{L}_{\text{loc}} \) represents the original localization loss function used in ImVoteNet\cite{imvotenet}. \( \mathcal{F}_{\text{text}} \) denotes the feature extracted by the text encoder in CLIP.

\textbf{Implementation Details}
Our model is a point cloud-only ImVoteNet \cite{imvotenet} +Clip architecture. The monocular depth estimation model used is ZoeDepth\cite{bhat2023zoedepth}, jointly trained on both indoor and outdoor scenes. In the pre-training phase, similar to ImVoteNet, we train for 180 epochs with an initial learning rate of 0.001. In the adaptation phase, we train for 100 epochs, reducing the learning rate to 0.0005.

For 2D voting, there are three types of cues: Geometric cues, Texture cues, and Semantic Cues. Unlike ImVoteNet, we retain geometric cues but remove texture cues. For Semantic cues, instead of using a one-hot class vector, we use pre-trained CLIP text encoder features, which is more suitable for an open-vocabulary setting.

\vspace{-0.2cm}
\section{Experiments}
\vspace{-0.2cm}

In this section, we compare our proposed \method{} with other baseline models. Our experimental setup is divided into two main stages: \textbf{Pretraining} and \textbf{Adaptation}. During the pretraining stage, the training data is pseudo 3D data, referring to pseudo 3D point clouds and their corresponding annotations (3D boxes). During the adaptation stage, we use ground truth point clouds and pseudo labels to minimize the domain gap. All experiments are conducted on two commonly used Object Detection datasets: SUNRGBD \cite{sunrgbd} and ScanNet \cite{scannet}. Additionally, we carry out comprehensive ablation studies to validate the effectiveness of our model's components and the data generation pipeline.

\subsection{Experimental Setup}

\textbf{2D Images Dataset:}
We select the LVIS \cite{lvis} dataset as our 2D image source for generating pseudo 3D data, utilizing 42,000 images provided in its training set, which spans 1,203 categories with rich and detailed annotations. 

\textbf{3D Point Clouds Dataset:} We select SUNRGBD \cite{sunrgbd} and ScanNet \cite{scannet} as our 3D point clouds datasets for adaptation and testing, SUNRGBD \cite{sunrgbd} and ScanNet \cite{scannet} encompass a diverse range of indoor environments and offer comprehensive annotations, including 2D and 3D bounding boxes for objects. We test on 20 common categories in both datasets.

\textbf{Evaluation Metrics:} We employ mean Average Precision (mAP) at an IoU threshold of 0.25 as our primary evaluation metric. This metric effectively balances precision and recall in assessing how well our models perform on selected datasets.
\begin{table}[h]
\centering
\caption{Results from the Pretraining stage comparison experiments on SUNRGBD and ScanNet, \method{} only require point clouds input.}
\label{tab:main_table1}
\resizebox{\textwidth}{!}
{%
\renewcommand{\arraystretch}{1.0}
\fontsize{6.5}{7}\selectfont
\begin{tabular}{ccccccc}
\toprule
\bfseries{Stage} & \textbf{Data Type} & \textbf{Method} & \textbf{Input} & \textbf{Training Strategy} & \textbf{\makecell{SUNRGBD\\ mAP@0.25}} & \textbf{\makecell{ScanNet\\ mAP@0.25}}\\
\midrule
 & & OV-VoteNet \cite{VoteNet} & Point Cloud & One-Stage & 5.18 & 5.86 \\
\textbf{Pre-}& \textbf{Pseudo} & OV-3DETR \cite{3DETR} & Point Cloud & One-Stage & 5.24 & 5.30\\
\textbf{training} & \textbf{Data} & OV-3DET \cite{ov3det} & Point Cloud + Image & Two-Stage & 5.47 & 5.69\\
 &  & \green \textbf{Ours} & \green \textbf{Point Cloud} &\green \textbf{One-Stage} & \green \textbf{12.61 \textcolor{rred}{$\uparrow$ 7.14}} & \green \textbf{12.64 \textcolor{rred}{$\uparrow$ 6.78}}\\
\bottomrule
\end{tabular}%
}
\end{table}

\begin{table}[h]
\centering
\caption{Results from the Adaptation stage comparison experiments on SUNRGBD and ScanNet}
\label{tab:main_table2}

\resizebox{\textwidth}{!}
{%
\renewcommand{\arraystretch}{1.0}
\fontsize{6}{6.5}\selectfont
\begin{tabular}{ccccccc}
\toprule
\bfseries{Stage}  & \textbf{Method} & \textbf{Input} & \textbf{Training Strategy} & \textbf{\makecell{SUNRGBD\\ mAP@0.25}} & \textbf{\makecell{ScanNet\\ mAP@0.25}}\\
\midrule
\textbf{Adap-}& \makecell{OV-3DET \cite{ov3det}\\ } & Point Cloud + Image & Two-Stage & 20.46 & 18.02\\
\textbf{tation}  & CoDA \cite{coda} & Point Cloud & One-Stage & --- & 19.32\\
 & \green \textbf{Ours} & \green \textbf{Point Cloud} &\green \textbf{One-Stage} & \green \textbf{22.53\textcolor{rred}{$\uparrow$ 2.07}} & \green \textbf{21.45\textcolor{rred}{$\uparrow$ 2.13}}\\
\bottomrule
\end{tabular}%
}
\vspace{-0.2cm}
\end{table}

\vspace{-0.2cm}
\subsection{Main Results}
\vspace{-0.2cm}
\textbf{Pretraining:} Due to the absence of existing baseline methods except OV-3DET \cite{ov3det}, we utilize CLIP \cite{CLIP} to make previous high-performance 3D detectors such as 3DETR \cite{3DETR} and VoteNet \cite{VoteNet}compatible with OV3Det. Specifically, to adapt traditional point cloud detectors for Open Vocabulary detection, we first extract geometric features from point clouds. Then, we integrate CLIP \cite{CLIP} for classification by converting these features for compatibility with CLIP \cite{CLIP} visual encoder and creating textual prompts for zero-shot classification. Finally, we compare the encoded prompts with the visual features to classify objects beyond the predefined categories. Therefore, these baselines are denoted as OV-VoteNet \cite{VoteNet}, OV-3DETR \cite{3DETR}.


\textbf{Adaptation:} To ensure a fair comparison with the current SOTA OV3Det methods, during the adaptation stage, all baselines use OV-3DET \cite{ov3det}'s approach to generating pseudo labels for ground truth point cloud data, which serve as training data for adaptation. In this stage, comparisons are made with CoDA \cite{coda} and OV-3DET \cite{ov3det}.

\vspace{-0.2cm}
\subsubsection{Pretraining $\rightarrow$ 3D Training Data Free OV-3Det}
\vspace{-0.2cm}

As shown in Table \ref{tab:main_table1}, training solely with pseudo 3D data generated by our method, \method{} improves mAP@0.25 by 7.14\% on SUNRGBD and 6.78\% on ScanNet over the best baseline. This achievement, made without using any 3D ground truth annotated data, demonstrates the high quality of our generated data and the effectiveness of using extensive 2D datasets to enhance Open World perception. Unlike OV-VoteNet, which lacks 2D image integration, our method's mAP@0.25 outperforms OV-VoteNet by 7.43\% and 6.78\% on the two datasets, proving the effectiveness of our multimodal approach even with only point cloud inputs. OV-3DET and \method{} visualization results are shown in Figure \ref{fig:kde}(b).

\vspace{-0.1cm}
\subsubsection{Adaptation $\rightarrow$ 3D Training Data Guided OV-3Det}
\vspace{-0.1cm}


Table \ref{tab:main_table2} shows original OV-3DET results in the first row. CoDA only compares with OV-3DET on ScanNet. Our experiments indicate that after pretraining with pseudo 3D data, \method{} outperforms the best baseline by 2.07\% on SUNRGBD and 2.13\% on ScanNet in mAP@0.25. This highlights the crucial role of pseudo 3D data in training and its effectiveness as data augmentation.

\vspace{-0.2cm}
\section{Ablation Study}
\vspace{-0.2cm}
\subsection{Ablation Study of 3D Data Revision}
To validate the effectiveness of enhancing pseudo 3D data quality, we conducted ablation experiments with the Rotation Correction Module and 3D Box Filtering Module. The 3D Box Filtering Module includes Train Phase Prior Size Filtering and Inference Phase Semantic Size Filtering. Table \ref{tab:ab1} shows the results: the baseline without any modules, adding Train Phase Prior Size Filtering improves mAP@0.25 by 1.65\% on SUNRGBD and 1.27\% on ScanNet. Adding the Rotation Correction Module improves by 1.3\% on SUNRGBD and 1.96\% on ScanNet. Combining both modules results in a 2.98\% improvement on SUNRGBD and 3.31\% on ScanNet. Adding Semantic Size Filtering during inference further increases mAP@0.25 by 4.26\% on SUNRGBD and 4.31\% on ScanNet. These results highlight the effectiveness of each module in improving data quality and OV3Det accuracy. \newline

\vspace{-0.5cm}
\begin{table}[H]
\renewcommand{\arraystretch}{1.2}
\centering
\scriptsize
\setlength{\tabcolsep}{3.2pt}
\caption{Results from the ablation study on the Rotation Correction Module and the 3D Box Filtering Module, conducted on SUNRGBD and ScanNet, are presented. The 3D Box Filtering Module is divided into two components: Train Phase Prior Size Filtering and Inference Phase Semantic Size Filtering.}
\label{tab:ab1}
\resizebox{0.8\textwidth}{!}{
\begin{tabular}{ccccccc}
\toprule
\bfseries{Stage}  & {\textbf{\makecell{Train Phase\\Prior Size}} } & {\textbf{\makecell{Rotation \\ Correction}} } & {\textbf{\makecell{Inference Phase\\Semantic Size}} } & \textbf{\makecell{SUNRGBD\\ mAP@0.25}}& \textbf{\makecell{ScanNet\\ mAP@0.25}}\\
\midrule
&\green\textcolor{rred}{\XSolidBrush} & \pink\textcolor{rred}{\XSolidBrush} & \purple\textcolor{rred}{\XSolidBrush} & 8.35 & 8.33  \\
\textbf{Pre-}&\green\textcolor{ggreen}{\checkmark} & \pink\textcolor{rred}{\XSolidBrush} & \purple\textcolor{rred}{\XSolidBrush} & 10.00 & 9.60  \\
\textbf{training}&\green\textcolor{rred}{\XSolidBrush} & \pink\textcolor{ggreen}{\checkmark} &\purple \textcolor{rred}{\XSolidBrush} & 9.65 & 10.29 \\
&\green \textcolor{ggreen}{\checkmark} & \pink\textcolor{ggreen}{\checkmark} & \purple\textcolor{rred}{\XSolidBrush} & 11.33 & 11.64 \\
&\green \textcolor{ggreen}{\checkmark} & \pink\textcolor{ggreen}{\checkmark} &\purple\textcolor{ggreen}{\checkmark} & \textbf{12.61} & \textbf{12.64} \\
\bottomrule
\end{tabular}
}
\end{table}

\begin{figure}[h]
    \centering
    \includegraphics[width=1\linewidth]{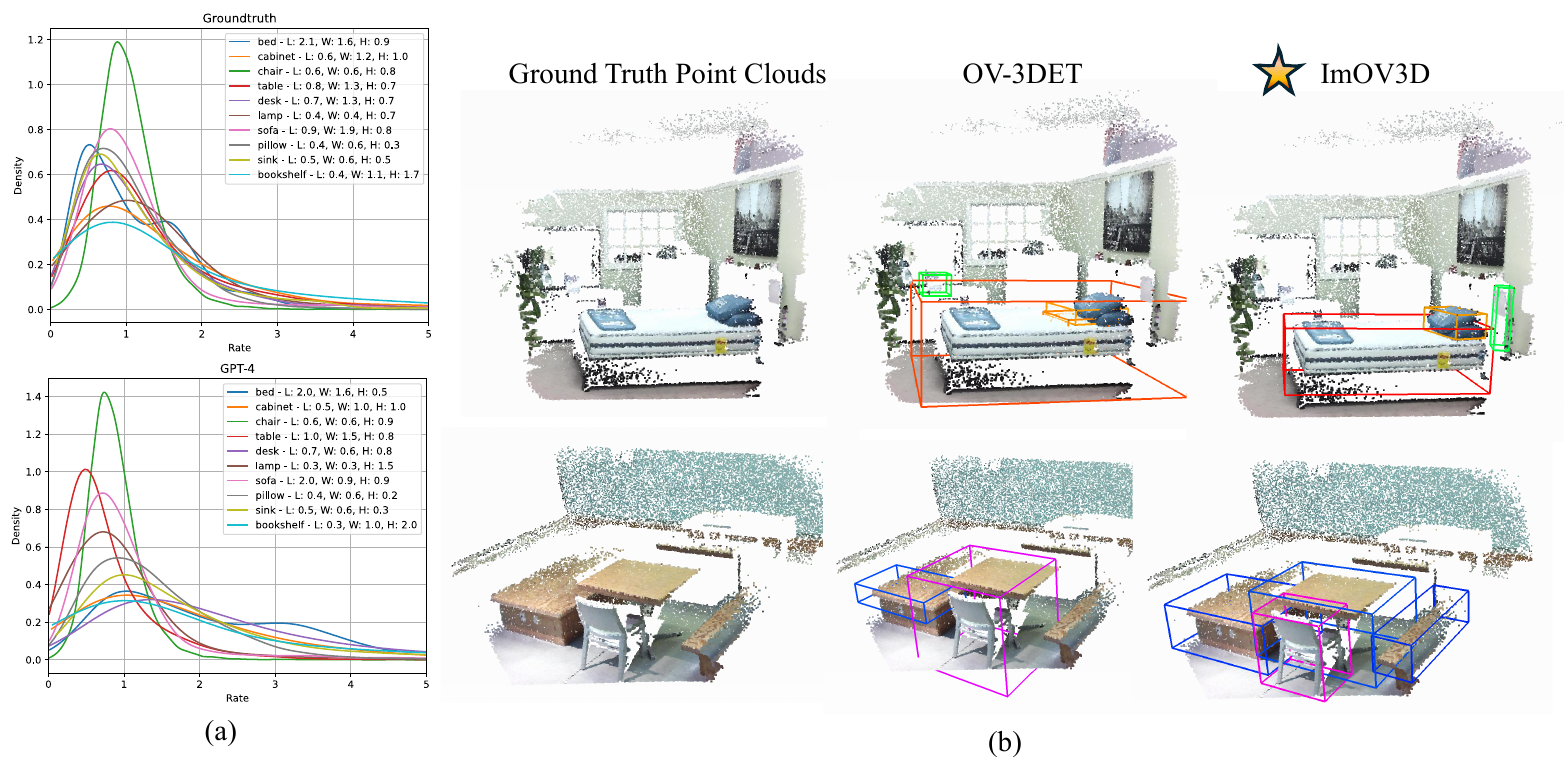}

    \caption{(a) KDE plots of volume ratios (\( \text{Ratio}_V \)) for top 10 classes in SUNRGBD validation set. (b) Visualization comparison of OV-3DET with ours in SUNRGBD.}
    \label{fig:kde}
\vspace{-0.5cm}
\end{figure}

We also discuss the efficiency of GPT-4 \cite{achiam2023gpt} in the 3D box filtering module using the SUNRGBD dataset \cite{sunrgbd}. For comparison, we select the top 10 classes with the most instances in the validation set. The volume ratio for these 10 classes is defined as \( \text{Ratio}_V = \frac{L \times W \times H}{L_{\text{GT/GPT}} \times W_{\text{GT/GPT}} \times H_{\text{GT/GPT}}} \). This ratio is an insightful metric for comparing the performance of the GPT-4 powered 3D box filter module to the ground truth (GT). A ratio close to 1 indicates high precision. We calculate \( \text{Ratio}_V \) for each instance and use Kernel Density Estimation (KDE) to analyze and plot the distributions of the volume ratios. Results are presented in Figure \ref{fig:kde}(a).

\subsection{Ablation Study of Depth and Pseudo Images}

\begin{wrapfigure}{r}{0.5\textwidth}
\begin{minipage}{0.5\textwidth}
\centering
\captionof{table}{The results from different types of 2D rendering images include depth maps and pseudo images.}
\label{tab:2dimagedatatyepe}
\renewcommand{\arraystretch}{1.2}
\setlength{\tabcolsep}{3.2pt}
\resizebox{\linewidth}{!}{%
\fontsize{7}{8}\selectfont
\begin{tabular}{cccc}
\toprule
\bfseries{Stage}  & {\textbf{\makecell{Rendered Images\\Data Types}} } & \textbf{\makecell{SUNRGBD\\ mAP@0.25}}& \textbf{\makecell{ScanNet\\ mAP@0.25}}\\
\midrule
\textbf{Pre-} &  \makecell{Depth Map}  & 4.38&4.47\\
\textbf{training} &\makecell{Pseudo Images} & \green \textbf{12.61} & \green \textbf{12.64}\\
\bottomrule
\end{tabular}%
}
\end{minipage}
\end{wrapfigure}

\begin{figure}[h]
    \centering
    \includegraphics[width=1\linewidth]{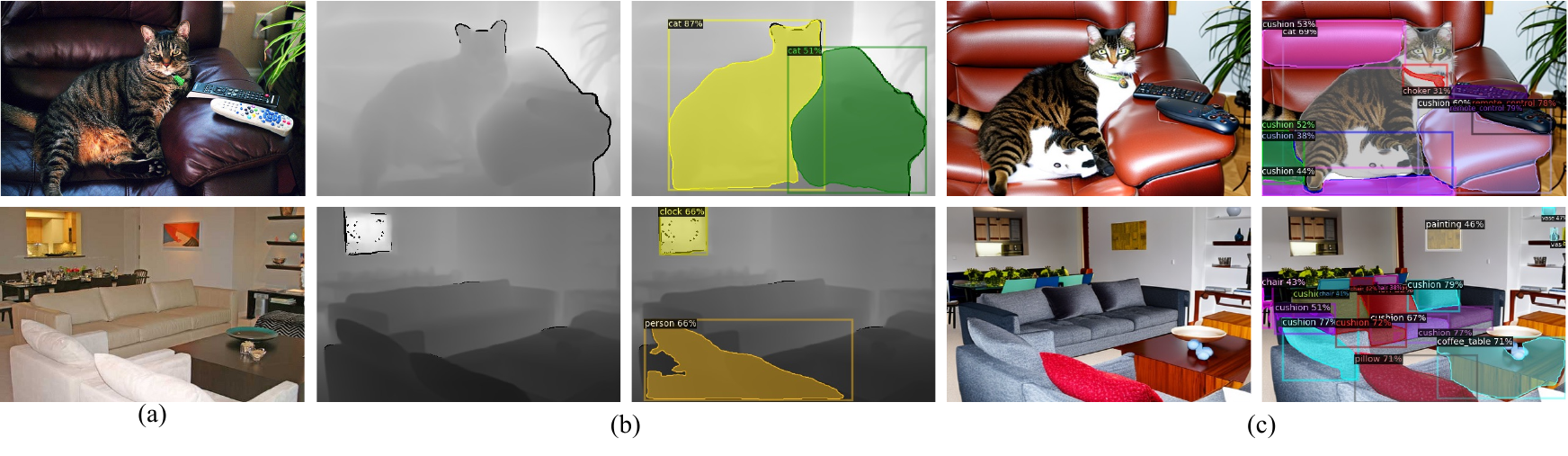}
    \caption{Qualitative results include (a) 2D RGB images, (b) 2D depth maps with 2D OVDetector annotations, and (c) pseudo images with annotations from a fine-tuned 2D detector. }
    \label{fig:depth_color}
\vspace{-0.3cm}
\end{figure}

To validate the effectiveness of pseudo images generated by ControlNet \cite{zhang2023adding}, we compare 2D depth maps from pseudo point clouds with pseudo images, shown in Figure \ref{fig:depth_color}. On the SUNRGBD dataset, mAP@0.25 increased from 4.38\% to 12.61\%, and on the ScanNet dataset, it rose from 4.47\% to 12.64\% (see Table \ref{tab:2dimagedatatyepe}). This shows that rich texture information in 2D images significantly enhances 3D detection performance.

\vspace{-0.1cm}
\subsection{Ablation Study of Data Volume}
\vspace{-0.2cm}

\begin{figure}[h]
    \centering
    \includegraphics[width=1\linewidth]{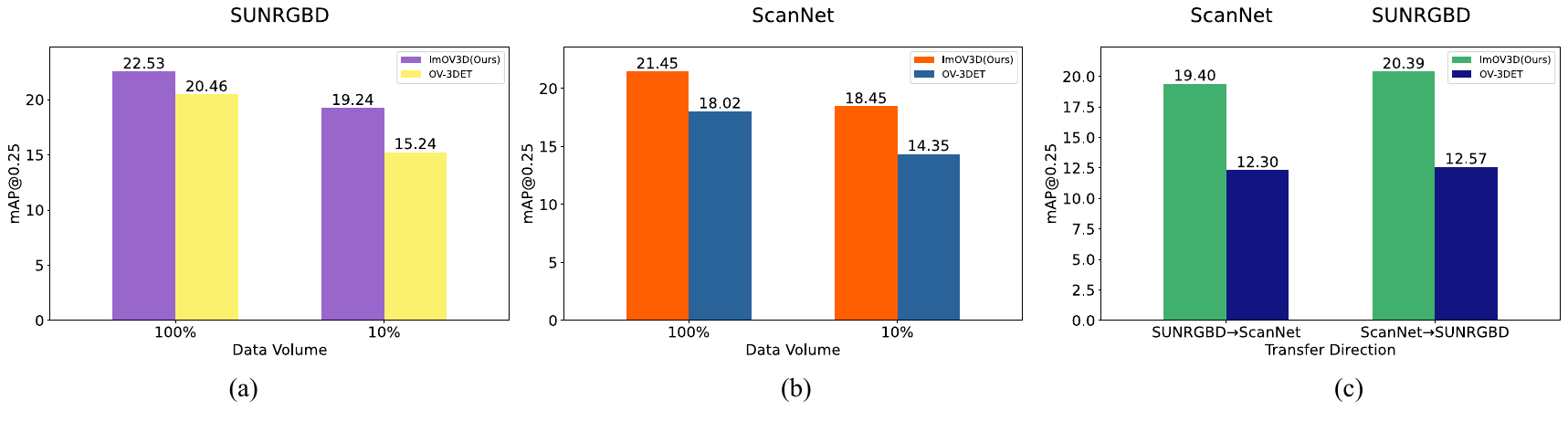}
    \caption{(a) and (b) show data volume ablation results. (c) illustrates transferability ablation results.}

    \label{fig:vol_trans}
\vspace{-0.5cm}
\end{figure}

Our method fine-tunes with limited real ground truth 3D point cloud data and pseudo 3D annotations. Using OV-3DET's code, we train with varying data volumes. With 10\% adaptation data, OV-3DET's mAP@0.25 on SUNRGBD drops from 20.46\% to 15.24\%, while ours drops from 22.53\% to 19.24\%. On ScanNet, OV-3DET's mAP@0.25 falls from 18.02\% to 14.35\%, and ours falls from 21.45\% to 18.45\% (Figure \ref{fig:vol_trans}(a)(b)). We observed a decrease in performance compared to using the full data set; however, our method was still able to maintain relatively high detection accuracy. This confirmed the robustness of our method and its adaptability to small datasets, enabling effective 3D Object Detection even under constrained data conditions. It also underscores the importance of developing models for OV3Det that are capable of learning from limited data and generalizing to a broader range of scenarios.
\vspace{-0.2cm}
\subsection{Analysis of Transferability}
\vspace{-0.2cm}
Traditional 3D detectors struggle with transferability due to training and testing class differences. We test \method{} on ScanNet and SUN RGB-D on the opposite datasets. The results, shown in Figure \ref{fig:vol_trans}(c), demonstrate that our model outperforms OV-3DET by 7.1\% on SUN RGB-D and 7.82\% on ScanNet. \method{} demonstrates superior transferability across domains despite the domain gap.

\vspace{-0.2cm}
\subsection{Analysis of Fine-tuned 2D Detector}
\vspace{-0.2cm}

\begin{table}[h]
\vspace{-0.5cm}
\centering
\caption{Comparison of fine-tuned 2D detector: Off-the-Shelf vs. Fine-Tuned Detic.}
\label{tab:fine-tuned 2D detector}
\resizebox{\textwidth}{!}
{%
\renewcommand{\arraystretch}{1.0}
\fontsize{6.5}{7}\selectfont
\begin{tabular}{ccccccc}
\toprule
\bfseries{Pretraining}  & {\textbf{Adaptation} } & \textbf{\makecell{SUNRGBD\\ mAP@0.25}}& \textbf{\makecell{ScanNet\\ mAP@0.25}}\\
\midrule
 - &  2D Off-the-shelf + 3D Adaptation  & 18.8 &18.96\\
Off-the-shelf + 3D Pretraining &  2D Off-the-shelf + 3D Adaptation  & 19.67 &19.25\\
\green2D Pretraining + 3D Pretraining & \green2D Adaptation + 3D Adaptation &\green\textbf{22.53} & \green\textbf{21.45}\\
\bottomrule
\end{tabular}%
}
\end{table}

To validate the benefits of fine-tuning Detic with pseudo images, we compare the off-the-shelf Detic to the fine-tuned version. The fine-tuned Detic shows clear advantages in handling pseudo images. On the SUNRGBD dataset, the mAP@0.25 increases from 19.67\% to 22.53\%, and on the ScanNet dataset, it rises from 19.25\% to 21.45\% (see Table \ref{tab:fine-tuned 2D detector}). These experiments were conducted under the adaptation setting, illustrating the model's ability to learn from and improve detection capabilities with not entirely real data. This not only confirms the efficacy of textured image but also highlights the importance of fine-tuning models to enhance their adaptability and accuracy.

\vspace{-0.2cm}

\vspace{-0.1cm}
\section{Conclusion and Limitation}
\vspace{-0.4cm}
 In conclusion, this paper introduce \method{}, a novel framework that tackles the scarcity of annotated 3D data in OV-3Det by harnessing the extensive availability of 2D images. The framework's key innovation lies in its flexible modality conversion, which integrates 2D annotations into 3D space, thereby minimizing the domain gap between training and testing data. Empirical results on two common datasets confirm ImOV3D's superiority over existing methods, even without ground truth 3D training data, and its significant performance boost with the addition of minimal real 3D data for fine-tuning. Our method's success showcases the potential of leveraging 2D images for enhancing 3D object detection, opening new avenues for future research in pseudo-multimodal data generation and its application in 3D detection methodologies. 

\textbf{Limitation:}Although our method has demonstrated the potential of 2D images in OV-3Det tasks, especially with the proposed pseudo multimodal representation, we need dense point clouds here to ensure that the rendered images can help improve performance. In the future, we will explore more generalized strategies.

\vspace{-0.1cm}
\section{Acknowledge}
\vspace{-0.4cm}
We would like to express our gratitude to Yuanchen Ju, Wenhao Chai, Macheng Shen, and Yang Cao for their insightful discussions and contributions.

\newpage
{
\small
\bibliographystyle{nips}
\bibliography{main}
}



\end{document}